\def\year{2019}\relax
\documentclass[letterpaper]{article} 
\usepackage{aaai19}  
\usepackage{times}  
\usepackage{helvet}  
\usepackage{courier}  
\usepackage{url}  
\usepackage{graphicx}  
\usepackage{amsmath}
\usepackage{amssymb}
\usepackage{amsthm}
\usepackage{amstext}
\usepackage{subfigure}
\usepackage{mathrsfs}
\usepackage{algorithm}
\usepackage{algpseudocode}
\theoremstyle{definition}
\newtheorem{definition}{Definition}

\begin{document}
%
\title{A Deep Reinforcement Learning Framework for Rebalancing Dockless Bike Sharing Systems \thanks{The work of Longbo Huang and Ling Pan was supported in part by the National Natural Science Foundation of China Grants 61672316, 61303195, the Tsinghua Initiative Research Grant, and the China Youth 1000-Talent Grant. 
Pingzhong Tang and Qingpeng Cai were supported in part by the National Natural Science Foundation of China Grant 61561146398, a China Youth 1000-talent program and an Alibaba Innovative Research program.}}
\author{
Ling Pan\textsuperscript{1}, Qingpeng Cai\textsuperscript{1}, Zhixuan Fang\textsuperscript{2}, Pingzhong Tang\textsuperscript{1}, Longbo Huang\textsuperscript{1}\\
\textsuperscript{1} IIIS, Tsinghua University\\
\textsuperscript{2} The Chinese University of Hong Kong\\
$\{$pl17, cqp14$\}$@mails.tsinghua.edu.cn, zxfang@ie.cuhk.edu.hk, $\{$kenshin, longbohuang$\}$@tsinghua.edu.cn
}	
\maketitle

\begin{abstract}
Bike sharing provides an environment-friendly way for traveling and is booming all over the world. Yet, due to the high similarity of user travel patterns, the bike imbalance problem constantly occurs, especially for dockless bike sharing systems, causing significant impact on service quality and company revenue. Thus, it has become a critical task for bike sharing operators to resolve such imbalance efficiently. In this paper, we propose a novel deep reinforcement learning framework for incentivizing users to rebalance such systems. We model the problem as a Markov decision process and take both spatial and temporal features into consideration. We develop a novel deep reinforcement learning algorithm called Hierarchical Reinforcement Pricing (HRP), which builds upon the Deep Deterministic Policy Gradient algorithm. Different from existing methods that often ignore spatial information and rely heavily on accurate prediction, HRP captures both spatial and temporal dependencies using a divide-and-conquer structure with an embedded localized module. We conduct extensive experiments to evaluate HRP, based on a dataset from Mobike, a major Chinese dockless bike sharing company. Results show that HRP performs close to the 24-timeslot look-ahead optimization, and outperforms state-of-the-art methods in both service level and bike distribution. It also transfers well when applied to unseen areas. 
\end{abstract} 

\section{Introduction}\label{section:intro}
Bike sharing, especially dockless bike sharing, is booming all over the world. 
For example, Mobike, a Chinese bike-sharing giant, has deployed over $7$ million bikes in China and abroad.
Being an environment-friendly approach, bike sharing provides people with a convenient way for commuting by sharing public bikes among users, and solves the ``last mile'' problem \cite{shaheen2010bikesharing}. 
Different from traditional docked bike sharing systems (BSS), e.g., Hubway, where bikes can only be rented and returned at fixed docking stations, users can access and park sharing bikes at any valid places. 
This relieves users' concerns about finding empty docks when they want to use bikes, or getting into fully occupied stations when they want to return them. 

However, due to similar travel patterns of most users, the rental mode of BSS leads to bike imbalance, especially during rush hours. For example, people mostly ride from home to work during morning peak hours. This results in very few bikes in residential areas, which in turn suppresses potential future demand, while subway stations and commercial areas are paralyzed due to the overwhelming number of shared bikes. 
This problem is further exaggerated for dockless BSS, due to unrestrained users' parking locations. 
This imbalance can cause severe problems not only to users and service providers, but also to cities. 
Therefore, it is crucial for bike sharing providers to rebalance bikes efficiently, so as to  serve users well and to avoid congesting city sidewalks and causing a bike mess. 

Bike rebalancing faces several challenges. 
First, it is a resource-constrained problem, as service providers often pose limited budgets for rebalancing the system. 
Naively spending the budget to increase the supply of bikes will not resolve the problem and is also not cost-efficient. Moreover, the number of bikes allowed is often capped due to regulation. 
Second, the problem is computationally intractable due to the large number of bikes and users. 
Third, the user demand is usually highly dynamic and changes both temporally and spatially. Fourth, if users are also involved in rebalancing bikes, the rebalancing strategy needs to efficiently utilize the budget and incentivize users to help, without knowing users' private costs.  

There have been a considerable set of recent results on bike rebalancing, which mainly focuses on two approaches, i.e., the vehicle-based approach \cite{o2015data,liu2016rebalancing,ghosh2016robust,li2018dynamic,ghosh2017incentivizing} and the user-based approach \cite{singla2015incentivizing,chemla2013self,fricker2016incentives}. 
The vehicle-based approach utilizes multiple trucks/bike-trailers to achieve repositioning by loading or unloading bikes in different regions. 
However, its rebalancing effect depends heavily on the accuracy of demand prediction. 
Also, it can be inflexible as it is hard to adjust the repositioning plan in real time to cater to the fluctuating demand. 
Additionally, due to the maintenance and traveling costs of trucks, as well as labor costs, the truck-based approach can deplete the limited budget rapidly. 
In contrast, the user-based approach offers a more economical and flexible way to rebalance the system, by offering users monetary incentives and alternative bike pick-up or drop-off locations. 
In this way, users are motivated to pick up or return bikes in neighboring regions rather than regions suffering from bike or dock shortage. 
However, existing user-based approaches often do not take the spatial information, including bike distribution and user distribution, into account in the incentivizing policy. 
Moreover, user related information, e.g., costs due to walking to another location,  is also often unknown.  

\begin{figure}
\centering
\includegraphics[width=0.4\textwidth]{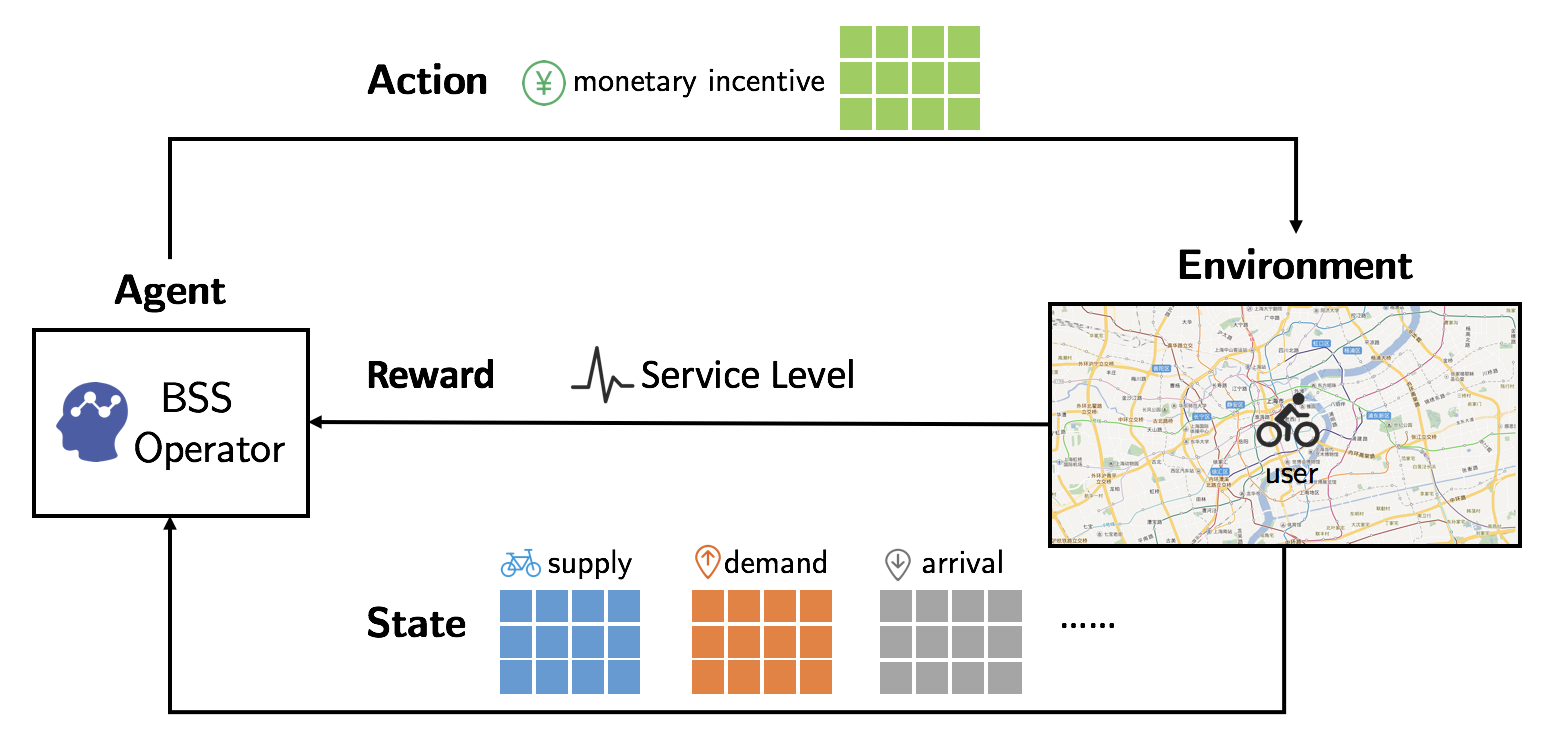}
\caption{An overview of the deep reinforcement learning framework for rebalancing dockless bikesharing systems.}
\label{fig: general_framework}
\end{figure}

In this paper, we propose a deep reinforcement learning framework for incentivizing users to rebalance dockeless BSS, as shown in Figure \ref{fig: general_framework}. 
Specifically, we view the problem as interactions between a bike sharing service operator and the environment, and formulate the problem as a Markov decision process (MDP). 
In this MDP, a state consists of supply, demand, arrival, and other related information, and each action corresponds to a set of monetary incentives 
for each region, to incentivize users to walk to nearby locations and use bikes there. The immediate reward is the number of satisfied user requests. 
Our objective is to maximize the long-term service level, i.e., the total number of satisfied requests, which is of highest interest for bike sharing service providers \cite{lin2011strategic}. 
Our approach falls under the topic of incentivizing strategies in multi-agent systems \cite{xue2016avicaching,tang2017reinforcement,cai2018reinforcement}.

To tackle our problem, we develop a novel deep reinforcement learning algorithm called the \emph{hierarchical reinforcement pricing} (HRP) algorithm. 
The HRP algorithm builds upon the Deep Deterministic Policy Gradient (DDPG) algorithm \cite{lillicrap2015continuous}, using the general hierarchical reinforcement learning framework \cite{dietterich2000hierarchical}. 
Our idea is to decompose the Q-value of the entire area of interest into multiple sub-Q-values of smaller regions. 
The decomposition enables an efficient searching for policies, as it addresses the complexity issue due to high-dimensional input space and temporal dependencies. 
In addition, the HRP algorithm also takes spatial dependencies into consideration and contains a localized module, in order to correct the bias in Q-value function estimation, introduced by  decomposition and correlations among sub-states and sub-actions. 
Doing so reduces the input space and decreases the training loss. We also show that the HRP algorithm improves convergence and achieves a better performance compared with existing algorithms. 

The main contributions of our paper are as follows: 
\begin{itemize}
\item We propose a novel spatial temporal bike rebalancing framework, and model the problem as a Markov decision process (MDP) that aims at maximizing the service level. 

\item We propose the \emph{hierarchical reinforcement pricing} (HRP) algorithm that decides how  to pay different users at each time, to incentivize them to help rebalance the system. 

\item We conduct extensive experiments using Mobike's dataset. Results show that HRP drastically outperforms state-of-the-art methods. 
We also validate the optimality of HRP by comparing with our proposed \emph{offline-optimal} algorithm. 
We further demonstrate HRP's generalization ability over different areas.
\end{itemize} 

\section{Related Work}
\textbf{Rebalancing approaches.} 
With the recent development of BSS, researchers have started to study operational issues leveraging big data \cite{liu2017functional,yang2016mobility,chen2016dynamic,li2015traffic}, among which rebalancing is one of the most important focuses. 
Rebalancing approaches can be classified into three categories. The first category adopts the truck-based approach which employs multiple trucks.
These methods can reposition bikes either in a static \cite{liu2016rebalancing} or dynamic \cite{ghosh2016robust} way, for docked BSS. 
The second category focuses on the use of bike-trailers \cite{o2015data,ghosh2017incentivizing,li2018dynamic}. 
The third category focuses on the user-based rebalancing approach \cite{singla2015incentivizing,chemla2013self,fricker2016incentives}. \\
\textbf{Reinforcement learning.} 
Deep Deterministic Policy Gradient algorithm (DDPG) \cite{lillicrap2015continuous} builds upon the Deterministic Policy Gradient algorithm \cite{silver2014deterministic}, using deep neural networks to approximate the action-value function for improving convergence. 
However, conventional reinforcement learning methods cannot scale up to problems with high-dimensional input spaces. 
Hierarchical reinforcement learning \cite{dayan1993feudal} decomposes a large problem into several smaller sub-problems learned by sub-agents, which is suitable for large-scale problems. 
Each sub-agent only focuses on learning sub-Q-values for its sub-MDP. 
Thus, the sub-agent can neglect part of the state which is irrelevant to its current decision \cite{andre2002state,van2017hybrid} to enable faster learning. 

 \section{Problem Definition}
In this section, we introduce and formalize the user-based bike rebalancing problem, i.e., by offering users monetary incentives to motivate them to help rebalance the system. 

Consider an area spatially divided into $n$ regions, i.e., $\{r_1, r_2, ..., r_n\}$. We discretize a day into $T$ timeslots with equal length, denoted by $\mathcal{T}=\{1, 2, ..., T\}$. 
Let $S_i(t)$ denote the supply in region $r_i$ at the beginning of timeslot $t\in\mathcal{T}$, i.e., the number of available bikes.
We also denote $\boldsymbol{S}(t) = (S_i(t), \forall\, i)$ as the vector of supply. 
The total user demand and bike arrival of region $r_i$ during the timeslot $t$ is denoted by $D_i(t)$ and $A_i(t)$ respectively. 
We similarly denote $\boldsymbol{D}(t) = (D_i(t), \forall\, i)$ and $\boldsymbol{A}(t) = (A_i(t), \forall\, i)$ as the vector of demand and the vector of arrival, respectively. 
Let $d_{ij}(t)$ denote the number of users intending to ride from region $r_i$ to region $r_j$ during the timeslot $t$.  

\noindent {\bf{Pricing algorithm.}} At each timeslot $t$, for a user who cannot find an available bike in his current region $r_i$,  
a pricing algorithm $\mathcal{A}$ suggests him alternate bikes in $r_i$'s neighboring regions, denoted by $N(r_i)$. 
Meanwhile, $\mathcal{A}$ also offers the user a price incentive $p_{ij}(t)$ (in the order of user arrivals), to motivate him to walk to neighboring region $r_j \in N(r_i)$ to pick up a bike $b_{hj}$, where $b_{hj}$ denotes the $h$-th bike in $r_j$.
$\mathcal{A}$ has a total rebalancing budget $B$. 
When the budget is completely depleted, $\mathcal{A}$ can no longer make further provision.

\noindent {\bf{User model.}} For each user $u_k$ in region $r_i$, if there are available bikes in the current region, he takes the nearest one.
Otherwise, there is a walking cost if he walks from $r_i$ to a neighboring region $r_j$ to pick up bikes. We denote the cost by $c_k(i, j, x)$, where $x$ is the walking distance from his location to the bike. We assume that it has the following form: 
\begin{eqnarray}
c_k(i,j,x)=
\begin{cases}
0 &           \text{$i$ equals to  $j$}\\
\alpha x^2  &        \text{$r_j$ is a neighboring region of $r_i$}\\
+\infty &           \text{else}\\
\end{cases}.
\label{user model}
\end{eqnarray}
This particular form of $c_k(i,j,x)$ is motivated by  a survey conducted in \cite{singla2015incentivizing}, where it is shown that user cost has a convex structure. 
Note that this cost is private to each user and the cost function is unknown to the service provider. 
For a user who cannot find an available bike in his current region $r_i$, if he receives an offer $(p_{ij}(t), b_{hj})$ and accepts it, he obtains a utility $p_{ij}(t)-c_k(i,j,x)$. 
Thus, a user will choose and pick up the bike $b_{hj}$ where he can obtain the maximum utility, and collect the price incentive $p_{ij}(t)$. If no offer leads to a nonnegative utility, the user will not accept any of them, resulting in an unsatisfied request. 

\noindent {\bf{Bike dynamics.}} Let $x_{ijl}(t)$ denote the number of users in $r_i$ riding to $r_l$ by taking a bike in $r_j$ at timeslot $t$. 
The dynamics of supply in each region $r_i$ can be expressed as: 
\begin{equation}
S_i(t+1) = S_i(t) - \sum_{j=1}^{n}\sum_{l=1}^{n} x_{jil}(t) + \sum_{m=1}^{n}\sum_{j=1}^{n} x_{mji}(t).
\label{eq: dynamics}
\end{equation}
The second and third terms in Eq. (\ref{eq: dynamics}) denote the numbers of  departing and arriving bikes in region $r_i$. 
Note that there is a travel time for bikes travel among regions. 

\noindent {\bf Objective.} Our goal is to develop an optimal pricing algorithm $\mathcal{A}$ to incentivize users in congested regions to pick up bikes in neighboring regions, so as to maximize the service level, i.e., the total number of satisfied requests, subject to the rebalancing budget $B$. 

\section{Hierarchical Reinforcement Pricing Algorithm}
\begin{algorithm*}
\caption{The Hierarchical Reinforcement Pricing (HRP) algorithm.}
\label{alg:RP}
\begin{algorithmic}
\Require Randomly initialize weights $\theta, \mu$ for the actor network $\pi_\theta(s)$ and the critic network $Q_\mu(s,a)$. 
\State Initialize target actor network $\pi'$ and target critic network $Q'$ with weights $\theta' \leftarrow \theta, \mu' \leftarrow \mu$
\State Initialize experience replay buffer $\mathcal{B}$ by filling with samples collected from the warm-up phase
\For {episode $= 1, ..., M$}
    \State Initialize a random process $\mathcal{N}$ to explore the action space, e.g. Gaussian noise, and receive initial state $s_1$
    \For {step $t = 1, ..., T$}
         \State $\triangleright$ \emph{\textbf{Explore and sample}}
    		\State Select and execute action $a_t = \pi_\theta(s_t) + \mathcal{N}_t$ ($\mathcal{N}_t$ sampled from $\mathcal{N}$), observe reward $R_t$ and next state $s_{t+1}$
    		\State Store transition $(s_t, a_t, R_t, s_{t+1})$ in experience replay buffer $\mathcal{B}$ //\emph{\textbf{update experience replay buffer $\mathcal{B}$}}
    		\State Sample a random minibatch of $N$ transitions $(s_i, a_i, R_i, s_{i+1})$ from $\mathcal{B}$
    		\State $\triangleright$ \emph{\textbf{Get current state-action pair's Q-value}}
    		\State Compute the decomposed Q-values $Q_{\mu_j}^j(s_{ji}, p_{ji})$ for each region $r_j$
    		\State Compute the bias-correction term $f_j(s_{ji}, NS(s_i, r_j), p_{ji})$ for each region $r_j$ by the localized module
    		\State Compute current state-action pair's Q-value $Q_{\mu}(s_i, a_i)$ according to Eq. (\ref{eq:Q-def})
    		\State $\triangleright$ \emph{\textbf{Get next state-action pair's Q-value}}
    		\State Get action for next state by actor network: $a'_{i+1} = \pi'_{\theta'} (s_{i+1})$
    		\State Compute the decomposed Q-values $Q_{\mu'_j}^j(s_{j(i+1)}, p'_{j(i+1)})$ for each region $r_j$
    		\State Compute the bias-correction term $f_j(s_{j(i+1)}, NS(s_{i+1}, r_j), p'_{j(i+1)})$ for each region $r_j$ by the localized module
    		\State Compute next state-action pair's Q-value $Q'_{\mu'}(s_{i+1}, a'_{i+1})$ according to Eq. (\ref{eq:Q-def})
    		\State $\triangleright$ \emph{\textbf{Update}}
    		\State Set $y_i = R_i + \gamma Q'_{\mu'}(s_{i+1}, a'_{i+1})$
    		\State Update the critic by minimizing the loss according to Eq. (\ref{eq:loss-def})
    		\State Update the actor using the sampled policy gradient according to Eq. (\ref{eq:J-def})
    		\State Update the target networks: $\theta' \leftarrow \tau \theta + (1 - \tau) \theta', \mu' \leftarrow \tau \mu + (1 - \tau) \mu'$
    \EndFor
\EndFor
\end{algorithmic}
\end{algorithm*}

In this section, we present our \emph{hierarchical reinforcement pricing} (HRP) algorithm that incentivizes users to rebalance the system efficiently. 

\subsection{MDP Formulation}
Our problem is an online learning problem, where an agent interacts with the environment. 
Therefore, it can be modeled as a Markov decision process (MDP) defined by a 5-tuple $(S, A, Pr, R, \gamma)$, where $S$ and $A$ denote the set of states and actions, $Pr$ the transition matrix, $R$ the immediate reward and $\gamma$ the discount factor. 
In our problem, at each timestep $t$,  the \textbf{state} $s_t=(\boldsymbol{S}(t), \boldsymbol{D}(t-1), \boldsymbol{A}(t-1), \boldsymbol{E}(t-1), RB(t), \boldsymbol{U}(t))$. 
Here, $\boldsymbol{S}(t)$ is the current supply for each region while $RB(t)$ is the current remaining budget. 
$\boldsymbol{D}(t-1)$, $\boldsymbol{A}(t-1)$ and $\boldsymbol{E}(t-1)$ are the demand, arrival and the expense in the last timestep for each region. 
$\boldsymbol{U}(t)$ represents the un-service rate for each region for a fixed number of past timesteps. 
The bike sharing service operator takes an \textbf{action $a_t = (p_{1t}, ..., p_{nt})$},  and receives an \textbf{immediate reward $R(s_t, a_t)$} which is the number of satisfied requests in the whole area $R$ at timestep $t$.
Specifically, $p_{it}$ represents the price for region $r_i$ at timestep $t$.\footnote{To reduce the complexity of the MDP, we employ the policy that the monetary incentives a user in region $r_i$ receives for picking up bikes in neighboring regions are the same. Thus, we use $p_{it}$ in place of $p_{ij}(t_k)$ for ease of notations.}
$Pr(s_{t+1} | s_t, a_t)$ represents the \textbf{transition probability} from state $s_t$ to state $s_{t+1}$ under action $a_t$.
The \textbf{policy function} $\pi_{\theta}(s_t)$ with the parameter $\theta$, maps the current state to a deterministic action. 
The overall \textbf{objective} is to find an optimal policy to maximize the \textbf{overall discounted rewards} from state $s_0$ following $\pi_{\theta}$, denoted by 
$J_{\pi_{\theta}} = \mathbb{E}[\sum_{k=0}^{\infty}{\gamma}^{k}R(a_k, s_k)|\pi_{\theta},s_0],$
where $\gamma \in [0, 1]$ denotes the \textbf{discount factor}. The Q-value of state $s_t$ and action $a_t$ under policy $\pi_{\theta}$ is denoted by $Q^{\pi_{\theta}}(s_t,a_t) =  \mathbb{E}[\sum_{k=t}^{\infty}{\gamma}^{k-t}R(s_k, a_k)|\pi_{\theta},s_t,a_t].$
Note that $J_{\pi_{\theta}}$ is a discounted version of the targeting service level objective, and will serve as a close approximation when $\gamma$ is close to $1$. 
Indeed, our experimental results show that with $\gamma=0.99$, our algorithm performs very close to the offline optimal of the service level objective, demonstrating the effectiveness of this approach. 

\subsection{The HRP Algorithm}
One of the key challenges in our problem, is that it has a continuous and high-dimensional action space that increases exponentially with the number of regions and suffers from the ``curse of dimensionality''. 
To tackle this problem, we propose the HRP algorithm, which is shown in Algorithm \ref{alg:RP}. Inspired by hierarchical reinforcement learning and DDPG, the HRP algorithm, which captures both temporal and spatial dependencies, is able to address the convergence issue of existing algorithms and improve performance. 

Specifically, we decompose the Q-value of the whole area into the sub-Q-value of each region. 
Then, the Q-value can be estimated by the additive combination of estimators of sub-Q-values according to $\sum_{j=1}^n Q_{\mu_j}^j(s_{jt}, p_{jt}),$ where $s_{jt}, p_{jt}$ denote the sub-state and sub-action of region $r_j$ at timestep $t$, and $\mu_j$ corresponds to the parameter of the estimator. 
For each timestep, the current state depends on previous states as people pick up and return bikes dynamically. 
To capture the sequential relationships exhibit in states, one idea is to train each sub-agent using Long Short-Term Memory (LSTM) \cite{hochreiter1997long} unit, which can capture complex sequential dependencies. 
However, LSTM maintains a complex architecture and needs to train a substantial number of parameters. 
Note that a key challenge in hierarchical reinforcement learning is to estimate accurate sub Q-values, which leads to an accurate overall Q-value estimation. 
Thus, we adopt the Gated Recurrent Unit (GRU) model \cite{cho2014learning}, a simplified variant of LSTM. Such a structure is more condensed and has fewer parameters, which enables a higher sample efficiency and avoids overfitting. 

However, applying a direct decomposition can lead to a large bias due to the dependence of each region and its neighbors. 
In our case, users may pick up bikes in neighboring regions besides their current regions, resulting in the fact that actions in different regions are coupled.
Therefore, we need to take this domain spatial feature into consideration. 
We tackle the bias of $Q_\mu(s_t,a_t)$ and $\sum_{j=1}^n Q_{\mu_j}^j(s_{jt}, p_{jt})$ by embedding the localized module to incorporate the spatial information. 
For each region $r_j$, the localized module (represented by $f_j$) takes not only the state $s_{jt}$ and action $p_{jt}$,  but also the states of its neighboring regions denoted by $NS(s_t, r_j)$ as inputs. 
In particular, $f_j$ is approximated by a neural network consisting of two fully-connected layers for bias correction. 
Thus, we estimate $Q_\mu(s_t, a_t)$ by: 
\begin{eqnarray} 
\sum_{j=1}^n Q_{\mu_j}^j(s_{jt}, p_{jt}) + f_j(s_{jt}, NS(s_t, r_j), p_{jt}). \label{eq:Q-def} 
\end{eqnarray}

The network architecture we propose to accelerate the search process in a large state space and a large action space is shown in Figure \ref{fig:framework}. Formally speaking, the key components of the HRP algorithm are described below: 

\begin{figure*}
\centering
\includegraphics[width=0.9\textwidth]{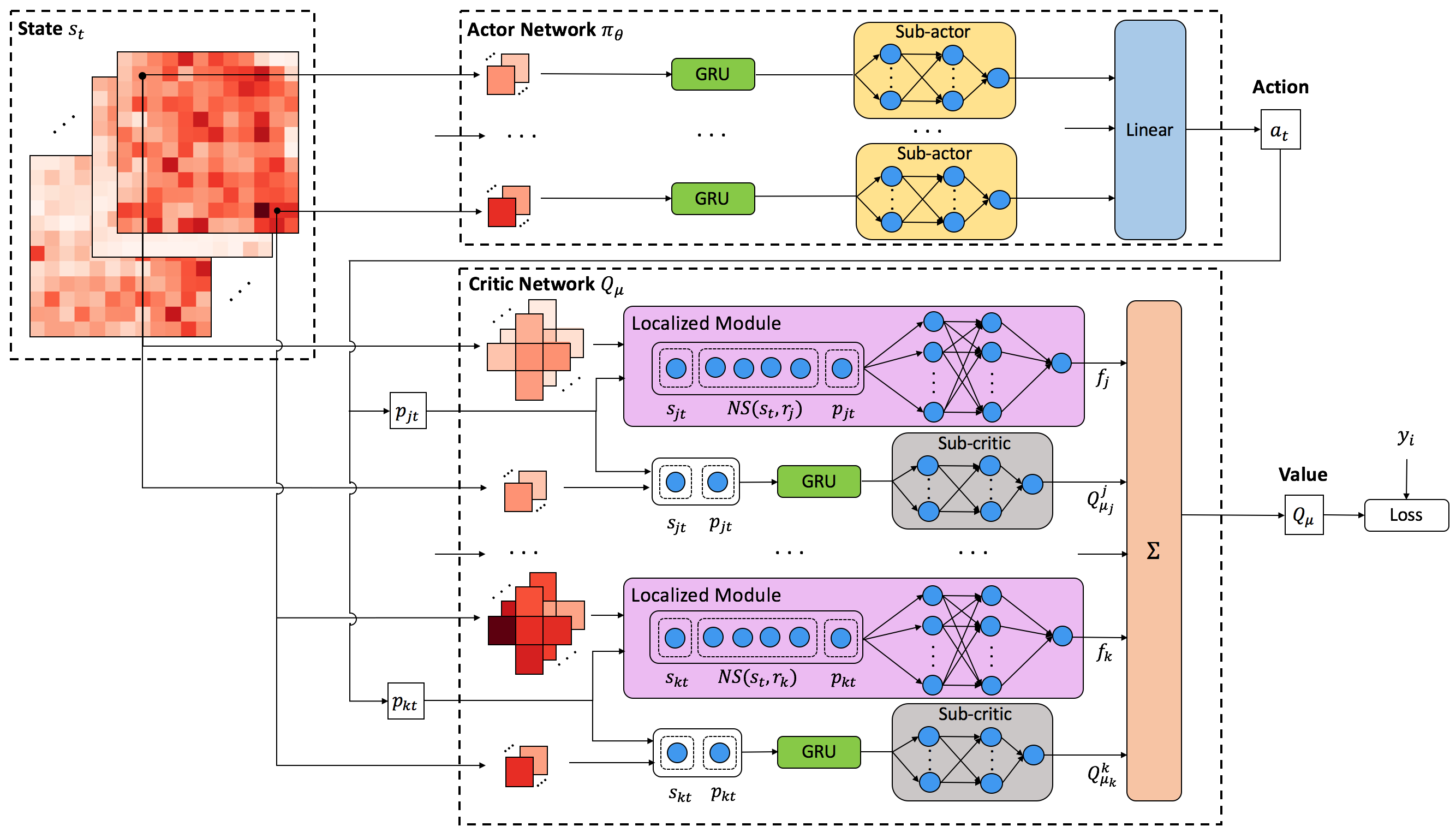}
\caption{The actor-critic framework of the HRP algorithm.}
\label{fig:framework}
\end{figure*}

\noindent \textbf{The actor:} The actor network represents the policy $\pi_\theta$ parameterized by $\theta$.
It maximizes $J_{\pi_\theta}$ using stochastic gradient ascent. In particular, the gradient of $J_{\pi_{\theta}}$ over $\theta$ is given by: 
\begin{eqnarray}
\nabla_{\theta}J_{\pi_{\theta}}=\mathbb{E}_{s\sim \rho_{\pi_{\theta}}}[\nabla_{\theta}\pi_{\theta}(s)\nabla_{a}Q_{\mu}(s,a)|_{a=\pi_{\theta}(s)}], \label{eq:J-def} 
\end{eqnarray}
where $\rho_{\pi_{\theta}}$ denotes the distribution of states.

\noindent \textbf{The critic:} The critic network takes the state $s_t$ and action $a_t$ as input, and outputs the action value. 
Specifically, the critic approximates the action-value function $Q^{\pi_{\theta}}(s,a)$ by minimizing the following loss \cite{lillicrap2015continuous}:
\begin{eqnarray}
L(Q^{\pi_\theta}) = \mathbb{E}_{s_t \sim \rho_{\pi}, a_t \sim \pi} [(Q_{\mu}(s_t, a_t) - y_t)^2], \label{eq:loss-def} 
\end{eqnarray}
where $y_t = R(s_t, a_t) + \gamma Q_{{\mu}^{'}}^{'}(s_{t+1}, \pi'_{\theta'} (s_{t+1}))$. 

\section{Experiments}

\subsection{Dataset}
\begin{figure}
    \centering
    \subfigure[Demand in 8 a.m.-9 a.m. 
    ]{
        \begin{minipage}[t]{0.45\linewidth}
            \centering
            \includegraphics[width=0.8\textwidth]{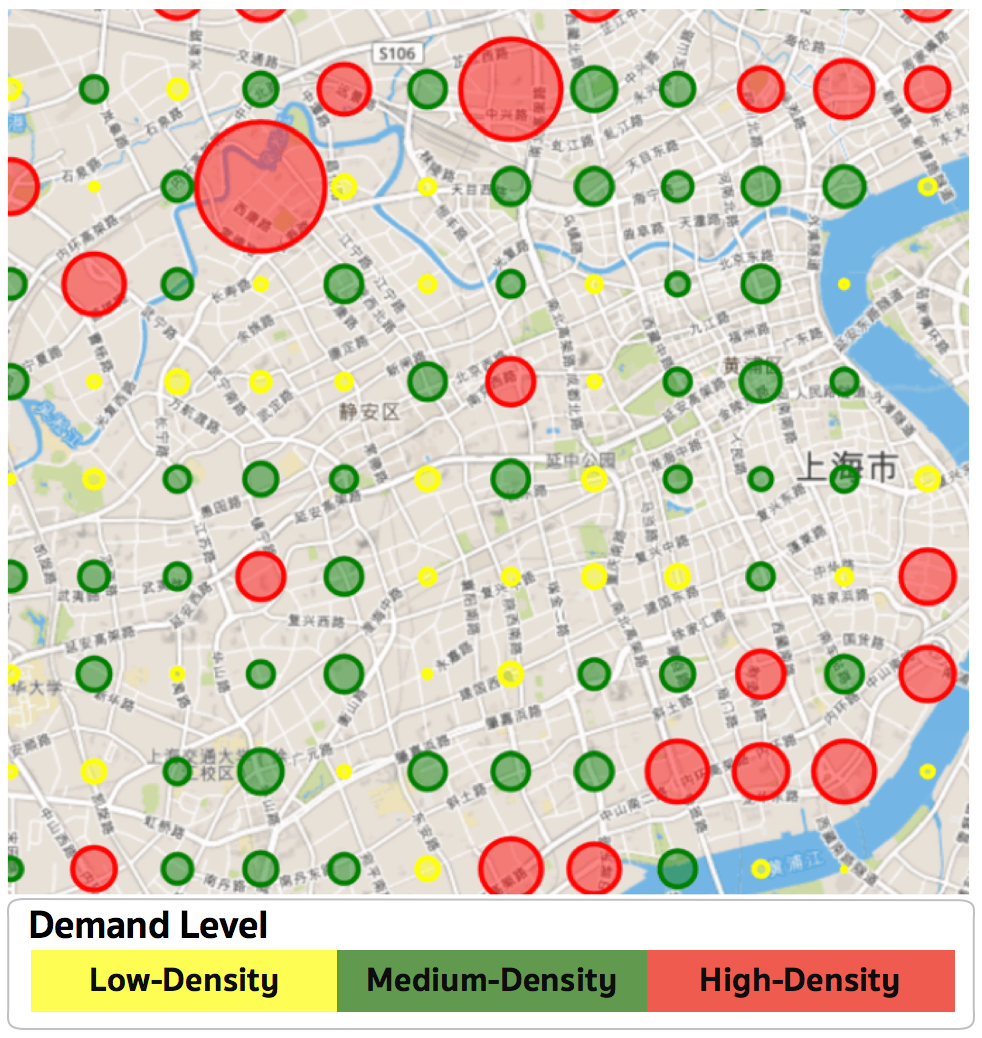}
        \end{minipage}
    }%
    \subfigure[Demand in 6 p.m.-7 p.m. 
    ]{
        \begin{minipage}[t]{0.45\linewidth}
            \centering
            \includegraphics[width=0.8\textwidth]{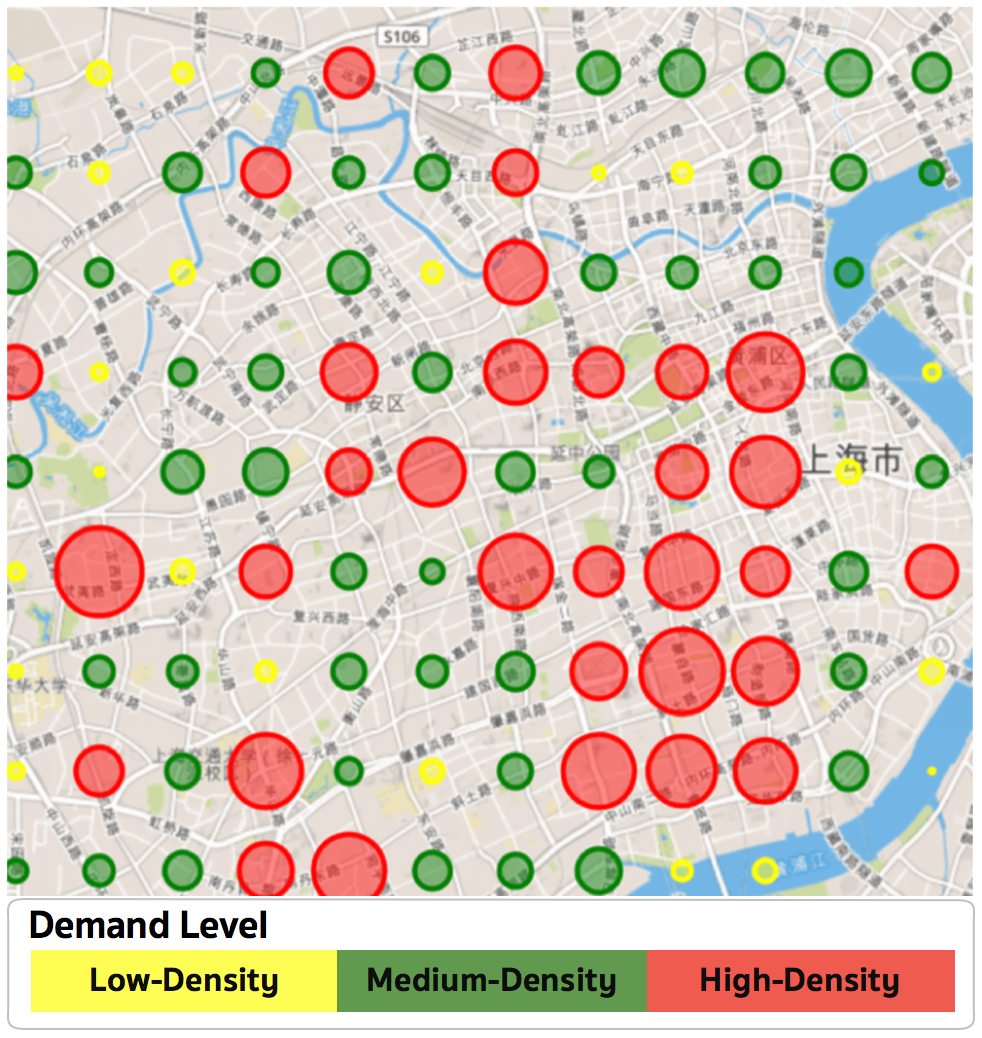}
        \end{minipage}
    }
    \caption{The temporal and spatial imbalance problem.}
    \label{fig: imbalance}
\end{figure}

We make use of a Mobike dataset consisting of users' trajectories from August $1$st to September $1$st in $2016$, in the city of Shanghai. 
Each data record contains the following information: order ID, bike ID, user ID, start time, start location (specified by longitude and latitude), end time, end location, and trace, with a total number of $102, 361$ orders. 
Figure \ref{fig: imbalance} shows the thermodynamic diagram of the dataset, where different colors of circles represent different demand levels with the radius representing the number of demand. As shown, there exists a significant temporal and spatial imbalance. 

\subsection{Experimental Setup}
According to Mobike's trajectory dataset, we obtain spatial and temporal information of users' requests and arrivals. 
We observe that the demand curve of different weekdays follows a very similar pattern according to the dataset, where the distribution during a day is bimodal with peaks corresponding to morning and evening rush hours. Thus, we aggregate all weekday demands to recover the total demand of a day. 
Each day consists of $24$ timeslots, and each timeslot is $1$ hour. 
We serve user requests every minute, where user requests are directly drawn from data while their starting locations (longitude and latitude) follow uniform distribution in their starting regions. 
To determine the initial distribution of bikes at the beginning of the day, we follow a common and practical way that most bike sharing operators (e.g. Mobike, Ofo) employ to redistribute bikes. The number of initial bikes in region $r_i$ is set as the product of the total supply and the ratio of the total demand in $r_i$ and the total demand for the whole area. Then, locations of initial bikes of each region are randomly drawn from data (from the locations of bikes in the corresponding region). 
Users respond to the system according to the defined user model, where the parameter $\alpha$ of the cost function as in Eq. (\ref{user model}) is selected so that the cost ranges in $[0 \ RMB, 5 \ RMB]$. 
This is chosen according to a survey conducted by \cite{singla2015incentivizing}, which shows that the cost ranges from $0$ to $2$ euros, and we converted it by the purchasing power of users.  
According to Mobike, the total number of orders in China is about $20$ million, with the total supply, i.e., number of bikes, in China to be $3.65$ million. 
Since we cannot obtain the total supply directly from data, 
we set the total supply as $O \times \frac{3.65}{20}$ ($O$ is the number of orders in our system). 
As the initial supply affects the inherent imbalance degree in Mobike's system, we vary this number in our experiments to evaluate this effect. 
We use the first-order approximation to compute the number of the unobserved lost demand \cite{o2015data}.

\subsubsection{Configurations of the HRP Algorithm}
For the state representation, we choose the fixed number of past timesteps of $\boldsymbol{U}(t)$ to be $8$. 
The comparison is fair as hyperparamters are the same for comparing reinforcement learning algorithms, following the configuration of \cite{lillicrap2015continuous}.
We train the algorithm for $100$ episodes in each setting, where each episode consists of $24N$ steps ($N$ denotes the number of days). After that, $20$ episodes are used for testing.

\subsection{Evaluation Metric}
We propose a metric called \emph{decreased un-service ratio} for performance evaluation.  
This choice can better characterize the improvement of the pricing algorithm over the original system (without monetary incentives) in the un-served events, compared with the service level \cite{singla2015incentivizing}.

\begin{definition}
\textbf{(Decreased un-service ratio)} 
The decreased un-service number of an algorithm $\mathcal{A}$ is the difference between the number of un-service events ($UN$) under Mobike and that of $\mathcal{A}$, i.e., 
$DUN(\mathcal{A}) = UN(\text{Mobike}) - UN(\mathcal{A})$. \footnote{Note that $UN(\text{Mobike})$ is computed by simulating Mobike's original system.} 
The decreased un-service ratio of $\mathcal{A}$ is defined as $DUR(\mathcal{A}) = \frac{DUN(\mathcal{A})}{UN(\text{Mobike})} \times 100\%$.
\label{eq:dun-def}
\end{definition}

\subsection{Baselines}
We compare HRP with the following baseline algorithms: 
\begin{itemize}
\item Randomized pricing algorithm (Random): assigns monetary incentives randomly under the budget constraint. 
\item OPT-FIX \cite{goldberg2001competitive}: an offline algorithm to maximize the acceptance rate with full a-priori user cost information.
\item DBP-UCB \cite{singla2015incentivizing}: a state-of-the-art method for user-based rebalancing which applied a multi-armed bandit framework. 
\item DDPG \cite{lillicrap2015continuous}
\item HRA \cite{van2017hybrid}: a reinforcement learning algorithm which directly employed reward decomposition. 
\item Offline-optimal: a pricing algorithm which maximizes the service level under the budget constraint with known user costs in advance, serving as an upper bound for any online pricing algorithms. 
To reduce the complexity of the formulation, we assume a trip finishes in the same timeslot (1 hour) it begins. \footnote{Otherwise, one has to set the timeslot as 1 minute, resulting in an integer linear program with too large complexity. Please note that the model and simulation do not rely on the assumption.}
In our comparison with the \emph{offline-optimal}, due to computational complexity, we assume that the costs of users walking from one region to another are the same for each timeslot $t$, defined as $c_k(i, j, x) = c_{ij}(t)$ if $r_j$ is the neighboring region of $r_i$, where $c_{ij}(t)$ is drawn independently from the empirical distribution of user costs for picking up bikes in $r_j$ instead of their current regions $r_i$ defined as Eq. (\ref{user model}) from timeslot $t$ of the dataset. 
Then, the problem can be formulated as the following integer linear program: \footnote{We solve the program with Gurobi.}
{\fontsize{9.0pt}{10.0pt} \selectfont
\begin{alignat}{1}
&\max\quad \sum_{t\in\mathcal{T}}\sum_{i,j,l \in [1,n]} x_{ijl}(t) \\
&\text{s.t.}\quad {x}_{ijl}(t)\in Z^{+}, \ \ \ \ \ \ \ \ \ \ \ \ \ \ \ \ \ \ \ \ \forall i, j, l, t\\
&\sum_{j \in [1, n]}{x}_{ijl}(t)\leq d_{il}(t), \ \ \ \ \ \ \ \ \ \ \ \ \ \ \forall i, l, t\label{eq:demand}\\ 
&\sum_{i, l \in [1, n]}{x}_{ijl}(t)\leq S_j(t), \ \ \ \ \ \ \ \ \ \ \ \ \ \forall j, t \label{eq:supply}\\
&\sum_{t\in\mathcal{T}}\sum_{i, j, l \in [1, n]}{x}_{ijl}(t)c_{ij}(t)\leq B \label{budget}\\ 
& S_i(t+1) =  S_i(t) - \sum_{j, l \in [1, n]} x_{jil}(t) +  \sum_{l, j \in [1, n]} x_{lji}(t), \, \forall i, t. \label{bike_supply_dynamics}
\end{alignat}
}
Constraints (\ref{eq:demand}) guarantee that for any timeslot, the number of users in $r_i$ heading for $r_l$ is no larger than the demand from $r_i$ to $r_l$. Constraints (\ref{eq:supply}) mean that the number of bikes picked up in $r_j$ does not exceed the bike supply in this region.
Constraint (\ref{budget}) ensures that the money the service provider spends is limited by $B$. 
Constraints (\ref{bike_supply_dynamics}) are for the dynamics of the supply of bikes.

\end{itemize}

\subsection{Performance Comparison}
We first compare HRP and reinforcement learning baselines to analyze the converging issue. 
Next, we compare HRP with varying budget and supply to evaluate its effectiveness in a day. 
Then, we analyze the long-term performance. 
Finally, we compare HRP with the \emph{offline-optimal} scheme and evaluate its generalization ability. 

\subsubsection{Training Loss} \label{subsection:loss}
Figure \ref{fig: loss_and_budget}(a) shows the training loss under HRP, HRA, and DDPG. 
As expected, DDPG diverges due to high-dimensional input spaces.
Notably, HRA diverges with an even  larger training loss. This is due to the impractical assumption of independence among sub-MDPs, which results in a biased estimator for the Q-value function. 
HRP outperforms the two and is able to converge. \footnote{The running time of HRP until convergence is less than 6 minutes on an NVIDIA GTX Titan-X 12GB graphics card.}

\subsubsection{Effect of varying budget constraints} 
\begin{figure}
    \centering
	    \subfigure[Training loss.]{
        \begin{minipage}[t]{0.48\linewidth}
            \centering
            \includegraphics[width=1.0\textwidth]{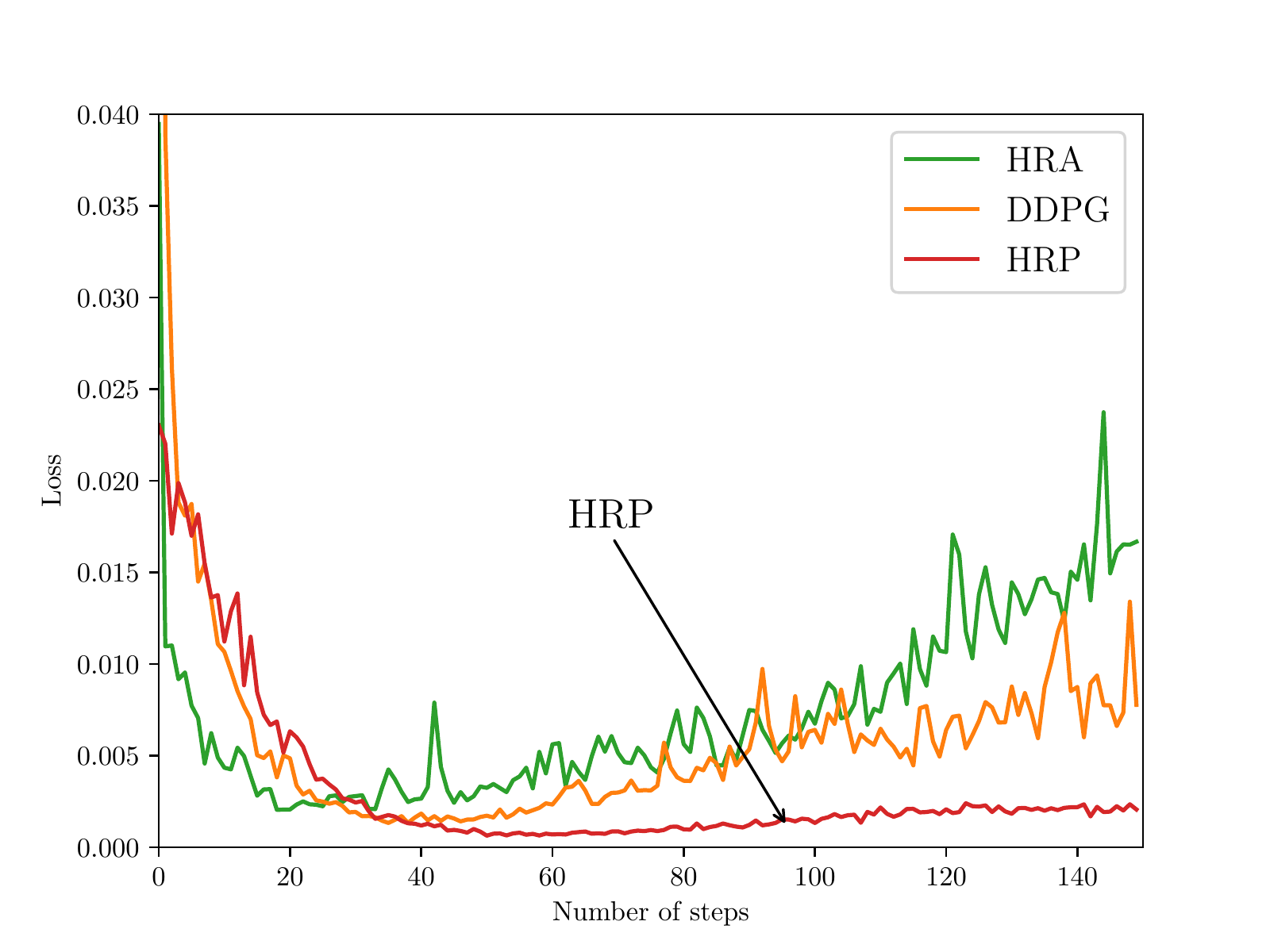}
        \end{minipage}
    }%
    \subfigure[Varying budget.]{
        \begin{minipage}[t]{0.48\linewidth}
            \centering
            \includegraphics[width=1.0\textwidth]{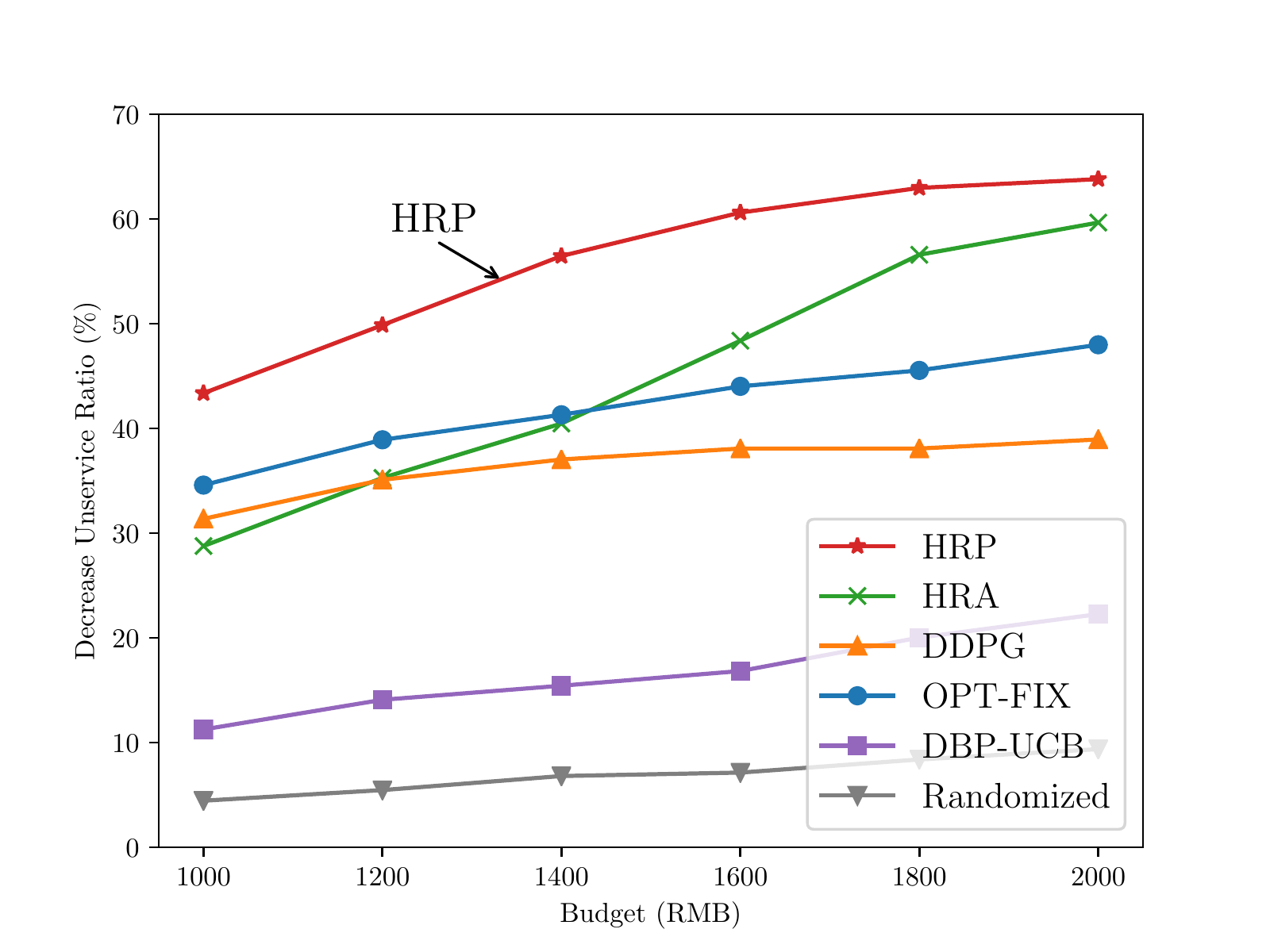}
        \end{minipage}
    }
    \caption{Comparison of training loss and varying budget.}    
    \label{fig: loss_and_budget}
\end{figure}

Figure \ref{fig: loss_and_budget}(b) shows the decreased un-service ratio of each algorithm under different budget constraints. 
DDPG performs similarly at different budget levels as it fails to learn to efficiently use the budget in such a large input space. 
HRA performs better than DDPG due to the decomposition. 
HRP decreases the un-service ratio by $43\%-63\%$, and outperforms other algorithms under all budget levels. 
The reason that OPT-FIX and DBP-UCB underperform HRP is that they do not consider spatial information, and focus solely on optimizing the acceptance rate.

\subsubsection{Achieving better bike distribution} \label{subsubsection:bike-dist}
To also analyze the rebalancing effect over a day, we measure how the distribution of bikes at the end of the day diverges from its value at the beginning of the day, using the Kullback-Leibler (KL) divergence measure \cite{kullback1951information}. 
We simulate Mobike's original system (without monetary incentives), 
and obtain its KL divergence level to be $0.554$. 
As seen in Table \ref{table: kld}, OPT-FIX obtains a bike distribution that is most different from the initial bike distribution. 
The reason can be that OPT-FIX is too aggressive in maximizing the acceptance rate, which can worsen the bike distribution.
HRP outperforms all existing algorithms with a KL divergence value of $0.548$, even smaller than Mobike's value. This demonstrates that HRP is able to improve the bike distribution at the end of a day.

\begin{table}
\centering
\begin{tabular}{c|c|c|c|c} 
\hline \hline
\textbf{HRP}  & \textbf{HRA} & \textbf{DBP-UCB} & \textbf{DDPG} & \textbf{OPT-FIX} \\
\hline
0.548 & 0.560 & 0.562 & 0.586 & 0.598\\
\hline \hline
\end{tabular}
\caption{KL divergence of user-based algorithms.}
\label{table: kld}
\end{table}

\subsubsection{Effect of varying supply} \label{subsubsection:varying_supply}
Besides varying the budget, it is also worth studying how HRP performs under different supply levels to evaluate its robustness. The results are shown in Figure \ref{fig: varying}(a). 
Intuitively, the problem is more challenging when supply is limited, which can lead to a large un-service rate. 
Random and DBP-UCB both perform poorly while the performance of OPT-FIX, DDPG, and HRA are almost the same. 
HRP performs significantly better than others, and achieves a $47\%-60\%$ decrement in the un-service ratio, 
demonstrating its robustness against different total supply. 

\subsubsection{Long-term performance} 
\begin{figure}
    \centering
       \subfigure[Varying supply.]{
        \begin{minipage}[t]{0.48\linewidth}
            \centering
            \includegraphics[width=1.0\textwidth]{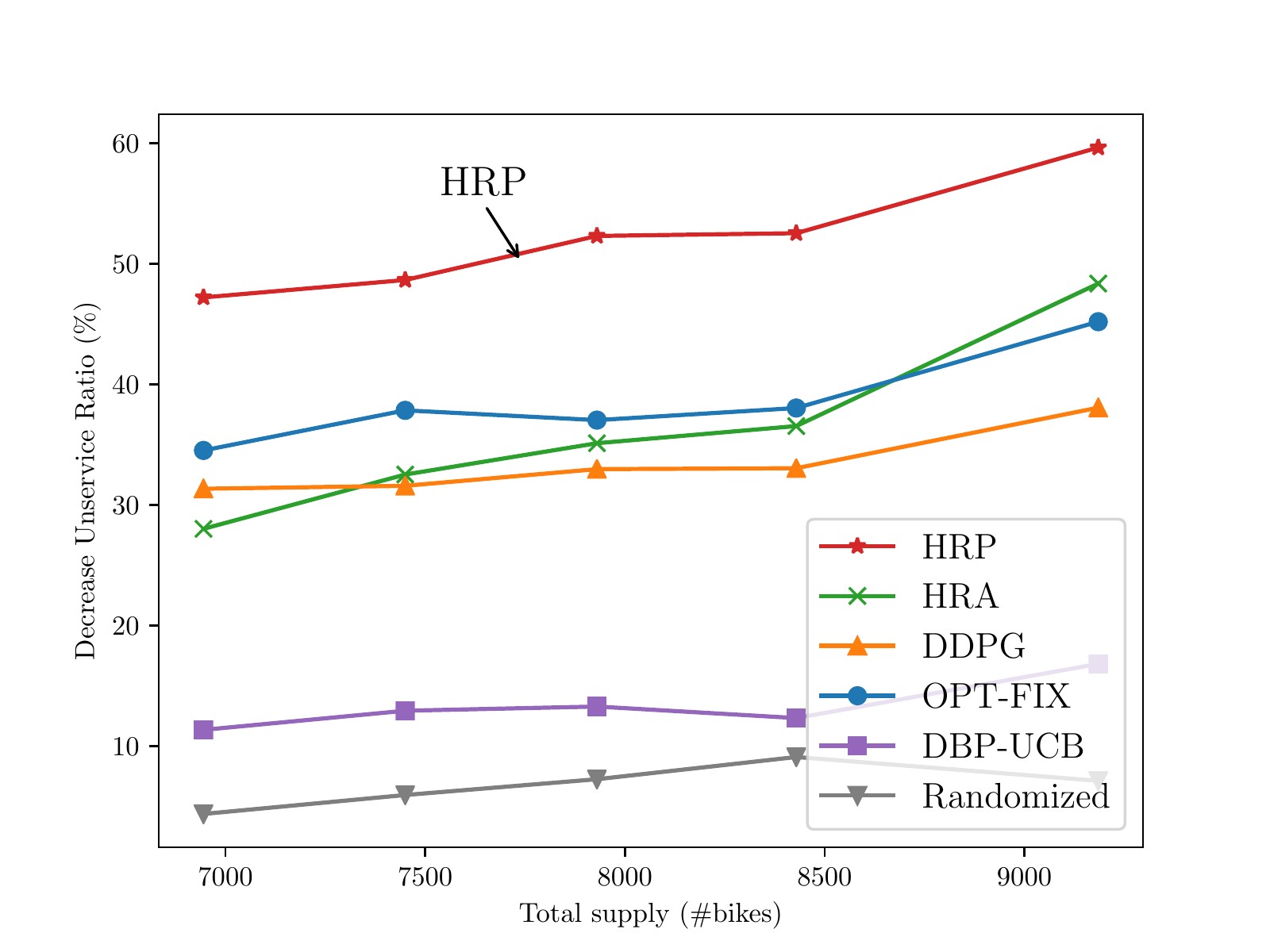}
        \end{minipage}
    }%
        \subfigure[Long-term.]{
        \begin{minipage}[t]{0.48\linewidth}
            \centering
            \includegraphics[width=1.0\textwidth]{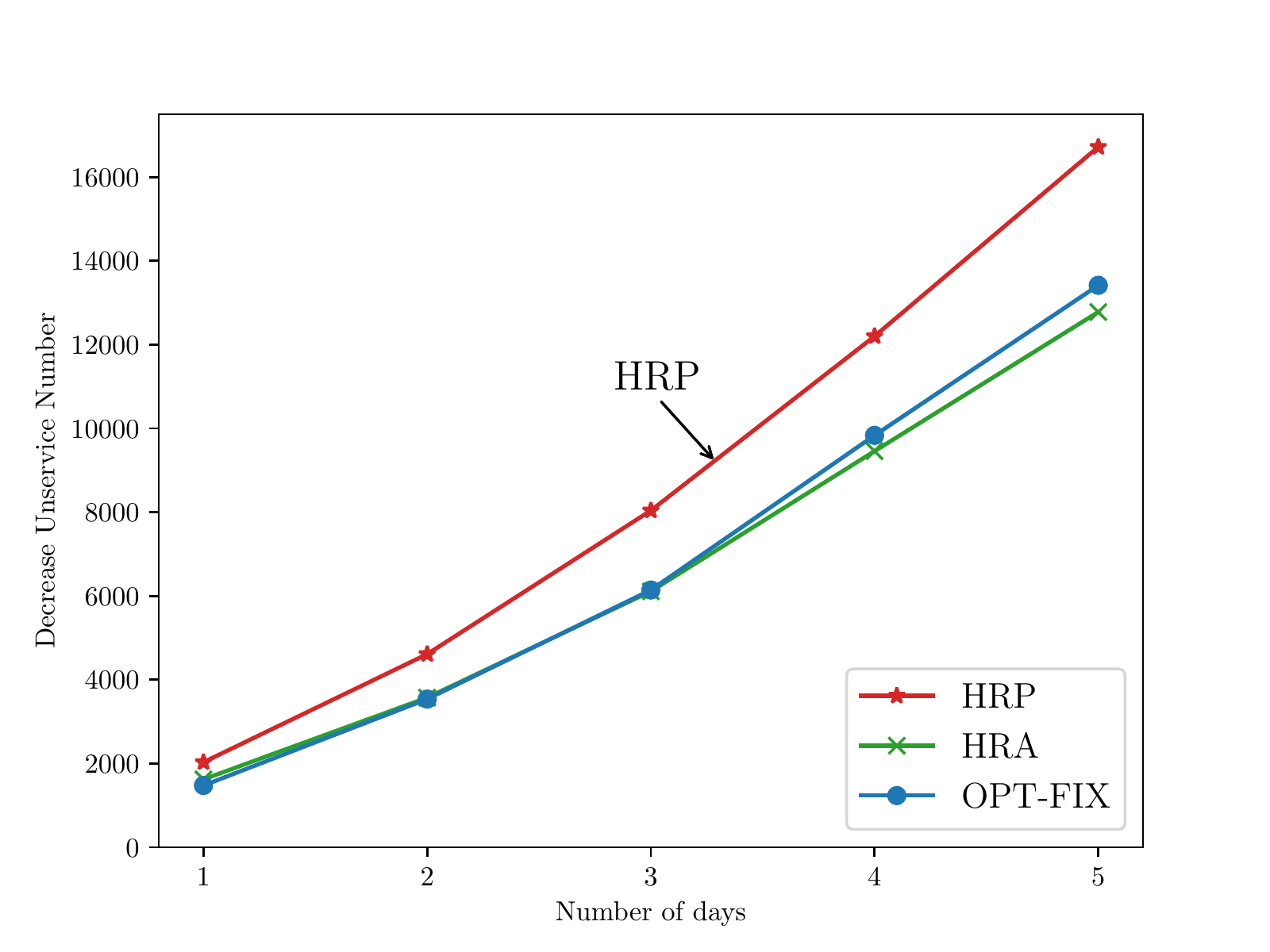}
        \end{minipage}
    }
    \caption{Comparison of varying supply and number of days.}    
    \label{fig: varying}
\end{figure}

Apart from analyzing one day's effect, we also evaluate the long-term performance by varying the number of days. 
We compare HRP with two most competitive algorithms, HRA and OPT-FIX, from $1$ day to $5$ days using decreased un-service number. 
As shown in Figure \ref{fig: varying}(b), HRP outperforms other algorithms. 
The performance gap also gets larger as the number of days increases. This is because HRP can achieve a better bike distribution as discussed in the previous section, and it learns to maximize the long-term reward. Readers please refer to the supplemental material for our detailed analysis of the result. 

\subsubsection{Optimality}
\begin{figure}
    \centering
    \subfigure[Optimality.]{
        \begin{minipage}[t]{0.48\linewidth}
            \centering
            \includegraphics[width=1.0\textwidth]{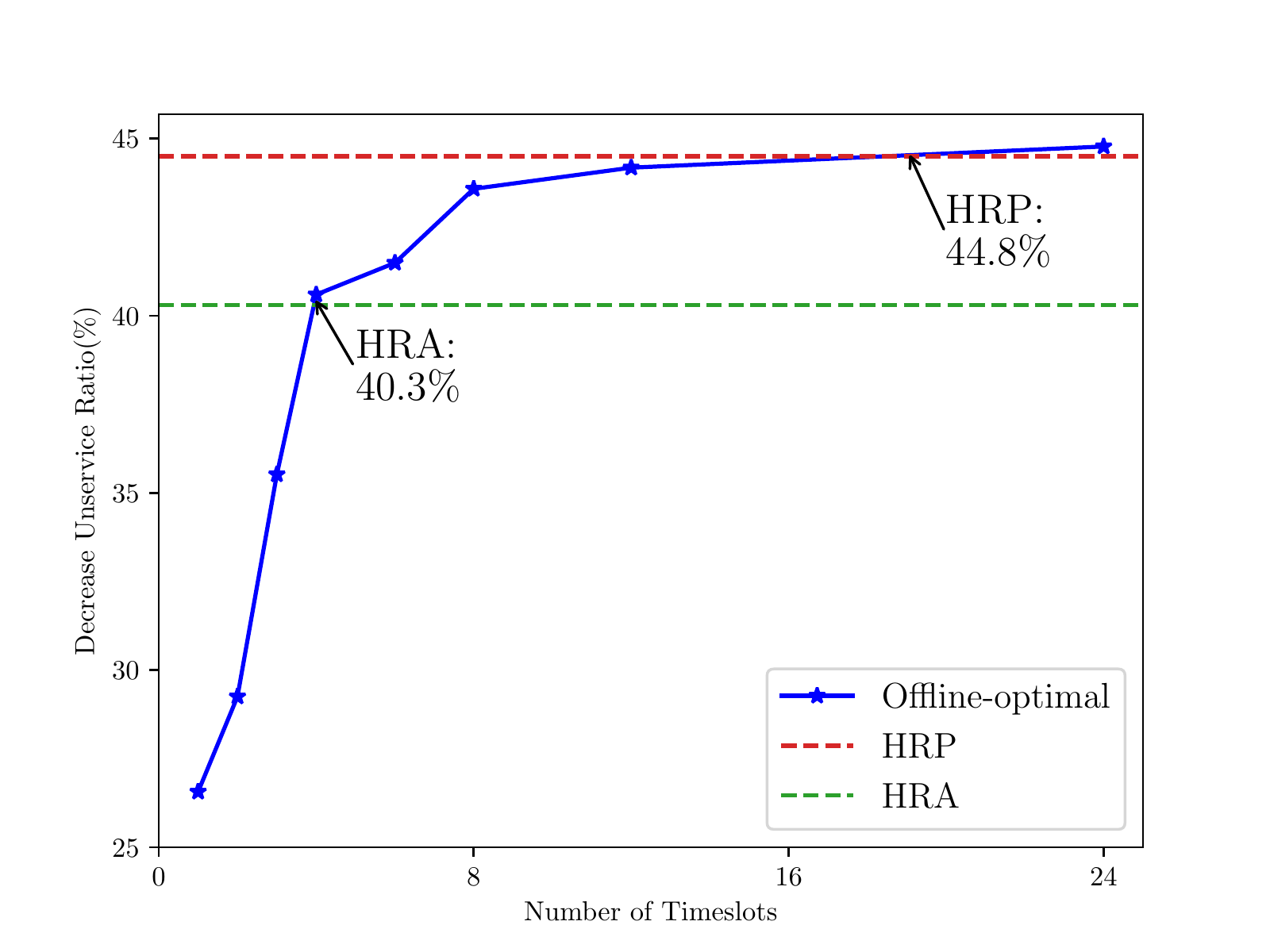}
            \label{fig:off_com}
        \end{minipage}
    }%
    \subfigure[Generalization.]{
        \begin{minipage}[t]{0.48\linewidth}
            \centering
            \includegraphics[width=1.0\textwidth]{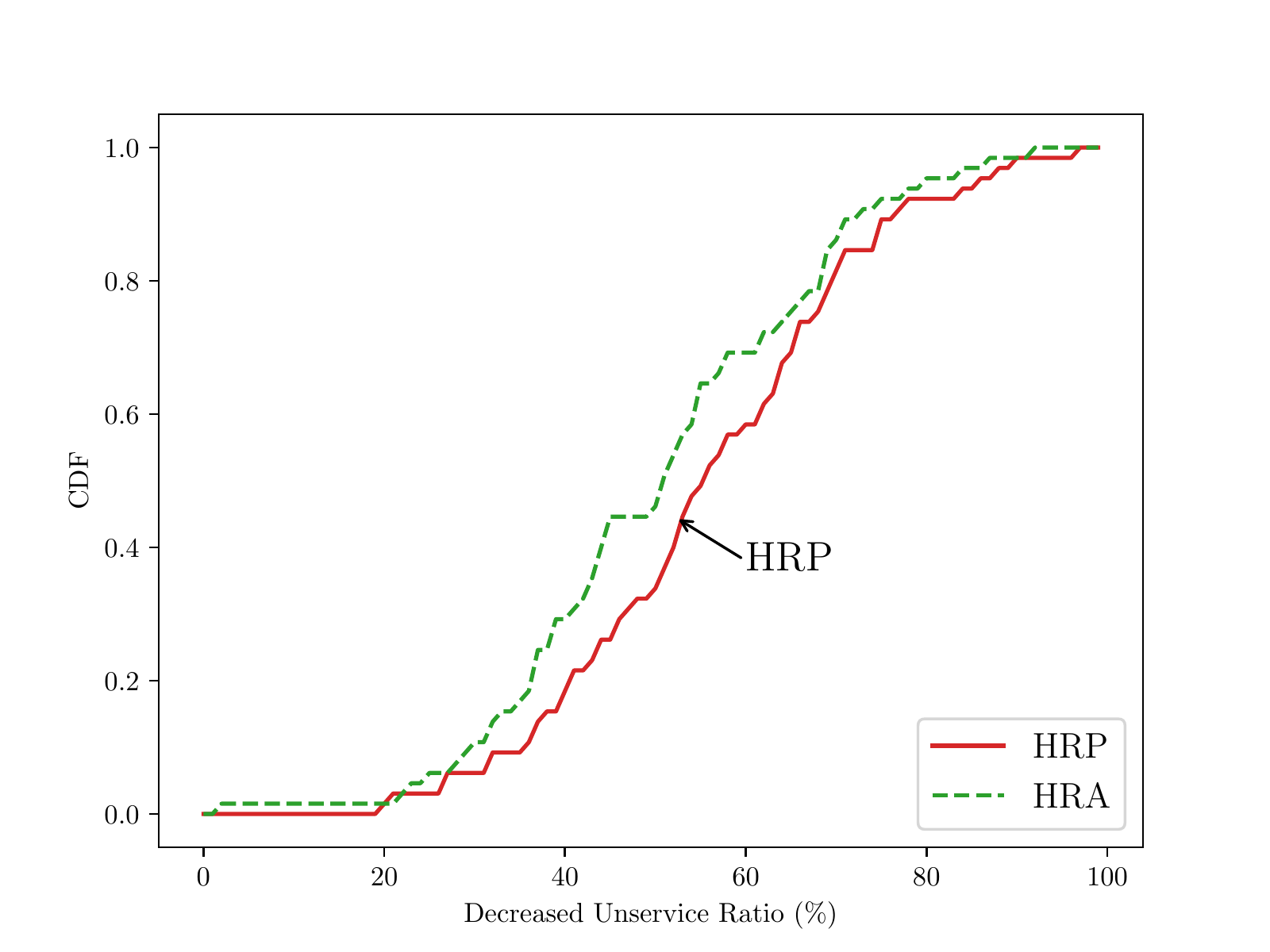}
            \label{fig:generalization}
        \end{minipage}
    }
    \caption{Optimality and generalization comparison.}
    \label{fig: opt_gen}
\end{figure}

We now provide comparison results  with the \emph{offline-optimal} scheme. 
To avoid computation and memory overhead, we conduct the comparison on a  smaller area consisting of $3\times3$ regions with highest request density to evaluate how well HRP can perform in a most congesting area. 
We compare both HRP and HRA with the \emph{offline-optimal}, by adjusting different values of timeslots $V$. 
The $V$-timeslot optimization means that for every $V$ timeslots, we optimize the offline program with $V$ horizons.  Intuitively, a larger $V$ leads to better performance. 
In this evaluation, the results are averaged over $400$ independent runs of each algorithm. 
Figure \ref{fig:off_com} demonstrates that the performance of HRP is very close to that of 24-timeslot optimization, while HRA only performs close to 4-timeslot optimization. 

\subsubsection{Generalization}
Now, we investigate whether HRP can transfer well, i.e., trained with certain areas but still performs well when applied to new ones. 
As the dataset only consists of trajectories in Shanghai, we divide the whole area into smaller areas where each consists of $3 \times 3$ regions to generate more areas. 
We train HRP and HRA in a certain area and then test them on other areas. 
Figure \ref{fig:generalization} shows the cumulative density function (CDF) of HRP and HRA on the decreased un-service ratio over all areas. 
HRP can achieve a $40\%-80\%$ un-service ratio decrement over $80\%$ areas in testing. 
This demonstrates that HRP generalizes well to new areas that it has never seen before. 
Note that the curve of HRP is strictly to the right of HRA, showing that HRP generalizes better than HRA. 

\section{Conclusion} 
We propose a deep reinforcement learning framework for incentivizing users to rebalance dockless bike sharing systems. 
We propose a novel deep reinforcement learning algorithm called \emph{hierarchical reinforcement pricing} (HRP). 
HRP maintains a divide-and-conquer structure that can handle complex temporal dependencies. 
It is also embedded with a localized module to better estimate the Q-value, which can capture spatial dependencies. 
We conduct extensive experiments for HRP based on real dataset from Mobike, a major Chinese dockless bike sharing company. 
Results show that HRP outperforms state-of-the-art methods. 
 
As for future work, one interesting extension is to deploy our algorithm in a real-world bike sharing system. 
To adapt to shifting environments across the year, one can first update the simulator with latest collected data, and then train HRP with the updated simulator offline every a fixed number of days. After training, one can adopt the new policy online. 
It is also promising to consider more factors in the user feedback model, e.g. travel distance and the time of day.

\fontsize{9.5pt}{10.0pt} \selectfont
\bibliographystyle{aaai}
\bibliography{aaai19}

\begin{thebibliography}{}

\bibitem[\protect\citeauthoryear{Andre and Russell}{2002}]{andre2002state}
Andre, D., and Russell, S.~J.
\newblock 2002.
\newblock State abstraction for programmable reinforcement learning agents.
\newblock In {\em AAAI/IAAI},  119--125.

\bibitem[\protect\citeauthoryear{Cai \bgroup et al\mbox.\egroup
  }{2018}]{cai2018reinforcement}
Cai, Q.; Filos-Ratsikas, A.; Tang, P.; and Zhang, Y.
\newblock 2018.
\newblock Reinforcement mechanism design for fraudulent behaviour in
  e-commerce.
\newblock In {\em Proceedings of the 32nd AAAI Conference on Artificial
  Intelligence}.

\bibitem[\protect\citeauthoryear{Chemla \bgroup et al\mbox.\egroup
  }{2013}]{chemla2013self}
Chemla, D.; Meunier, F.; Pradeau, T.; Calvo, R.~W.; and Yahiaoui, H.
\newblock 2013.
\newblock Self-service bike sharing systems: simulation, repositioning,
  pricing.

\bibitem[\protect\citeauthoryear{Chen \bgroup et al\mbox.\egroup
  }{2016}]{chen2016dynamic}
Chen, L.; Zhang, D.; Wang, L.; Yang, D.; Ma, X.; Li, S.; Wu, Z.; Pan, G.;
  Nguyen, T.-M.-T.; and Jakubowicz, J.
\newblock 2016.
\newblock Dynamic cluster-based over-demand prediction in bike sharing systems.
\newblock In {\em Proceedings of the 2016 ACM International Joint Conference on
  Pervasive and Ubiquitous Computing},  841--852.
\newblock ACM.

\bibitem[\protect\citeauthoryear{Cho \bgroup et al\mbox.\egroup
  }{2014}]{cho2014learning}
Cho, K.; Van~Merri{\"e}nboer, B.; Gulcehre, C.; Bahdanau, D.; Bougares, F.;
  Schwenk, H.; and Bengio, Y.
\newblock 2014.
\newblock Learning phrase representations using rnn encoder-decoder for
  statistical machine translation.
\newblock In {\em Proceedings of the 2014 Conference on Empirical Methods in
  Natural Language Processing (EMNLP)},  1724--1734.

\bibitem[\protect\citeauthoryear{Dayan and Hinton}{1993}]{dayan1993feudal}
Dayan, P., and Hinton, G.~E.
\newblock 1993.
\newblock Feudal reinforcement learning.
\newblock In {\em Advances in neural information processing systems},
  271--278.

\bibitem[\protect\citeauthoryear{Dietterich}{2000}]{dietterich2000hierarchical}
Dietterich, T.~G.
\newblock 2000.
\newblock Hierarchical reinforcement learning with the maxq value function
  decomposition.
\newblock {\em Journal of Artificial Intelligence Research} 13:227--303.

\bibitem[\protect\citeauthoryear{Fricker and
  Gast}{2016}]{fricker2016incentives}
Fricker, C., and Gast, N.
\newblock 2016.
\newblock Incentives and redistribution in homogeneous bike-sharing systems
  with stations of finite capacity.
\newblock {\em Euro journal on transportation and logistics} 5(3):261--291.

\bibitem[\protect\citeauthoryear{Ghosh and
  Varakantham}{2017}]{ghosh2017incentivizing}
Ghosh, S., and Varakantham, P.
\newblock 2017.
\newblock Incentivizing the use of bike trailers for dynamic repositioning in
  bike sharing systems.

\bibitem[\protect\citeauthoryear{Ghosh, Trick, and
  Varakantham}{2016}]{ghosh2016robust}
Ghosh, S.; Trick, M.; and Varakantham, P.
\newblock 2016.
\newblock Robust repositioning to counter unpredictable demand in bike sharing
  systems.
\newblock In {\em Proceedings of the Twenty-Fifth International Joint
  Conference on Artificial Intelligence}, IJCAI'16,  3096--3102.
\newblock AAAI Press.

\bibitem[\protect\citeauthoryear{Goldberg, Hartline, and
  Wright}{2001}]{goldberg2001competitive}
Goldberg, A.~V.; Hartline, J.~D.; and Wright, A.
\newblock 2001.
\newblock Competitive auctions and digital goods.
\newblock In {\em Proceedings of the twelfth annual ACM-SIAM symposium on
  Discrete algorithms},  735--744.
\newblock Society for Industrial and Applied Mathematics.

\bibitem[\protect\citeauthoryear{Hochreiter and
  Schmidhuber}{1997}]{hochreiter1997long}
Hochreiter, S., and Schmidhuber, J.
\newblock 1997.
\newblock Long short-term memory.
\newblock {\em Neural computation} 9(8):1735--1780.

\bibitem[\protect\citeauthoryear{Kullback and
  Leibler}{1951}]{kullback1951information}
Kullback, S., and Leibler, R.~A.
\newblock 1951.
\newblock On information and sufficiency.
\newblock {\em The annals of mathematical statistics} 22(1):79--86.

\bibitem[\protect\citeauthoryear{Li \bgroup et al\mbox.\egroup
  }{2015}]{li2015traffic}
Li, Y.; Zheng, Y.; Zhang, H.; and Chen, L.
\newblock 2015.
\newblock Traffic prediction in a bike-sharing system.
\newblock In {\em Proceedings of the 23rd SIGSPATIAL International Conference
  on Advances in Geographic Information Systems}, ~33.
\newblock ACM.

\bibitem[\protect\citeauthoryear{Li, Zheng, and Yang}{2018}]{li2018dynamic}
Li, Y.; Zheng, Y.; and Yang, Q.
\newblock 2018.
\newblock Dynamic bike reposition: A spatio-temporal reinforcement learning
  approach.
\newblock In {\em Proceedings of the 24th ACM SIGKDD International Conference
  on Knowledge Discovery and Data Mining},  1724--1733.
\newblock ACM.

\bibitem[\protect\citeauthoryear{Lillicrap \bgroup et al\mbox.\egroup
  }{2015}]{lillicrap2015continuous}
Lillicrap, T.~P.; Hunt, J.~J.; Pritzel, A.; Heess, N.; Erez, T.; Tassa, Y.;
  Silver, D.; and Wierstra, D.
\newblock 2015.
\newblock Continuous control with deep reinforcement learning.
\newblock {\em arXiv preprint arXiv:1509.02971}.

\bibitem[\protect\citeauthoryear{Lin and Yang}{2011}]{lin2011strategic}
Lin, J.-R., and Yang, T.-H.
\newblock 2011.
\newblock Strategic design of public bicycle sharing systems with service level
  constraints.
\newblock {\em Transportation research part E: logistics and transportation
  review} 47(2):284--294.

\bibitem[\protect\citeauthoryear{Liu \bgroup et al\mbox.\egroup
  }{2016}]{liu2016rebalancing}
Liu, J.; Sun, L.; Chen, W.; and Xiong, H.
\newblock 2016.
\newblock Rebalancing bike sharing systems: A multi-source data smart
  optimization.
\newblock In {\em Proceedings of the 22nd ACM SIGKDD International Conference
  on Knowledge Discovery and Data Mining},  1005--1014.
\newblock ACM.

\bibitem[\protect\citeauthoryear{Liu \bgroup et al\mbox.\egroup
  }{2017}]{liu2017functional}
Liu, J.; Sun, L.; Li, Q.; Ming, J.; Liu, Y.; and Xiong, H.
\newblock 2017.
\newblock Functional zone based hierarchical demand prediction for bike system
  expansion.
\newblock In {\em Proceedings of the 23rd ACM SIGKDD International Conference
  on Knowledge Discovery and Data Mining},  957--966.
\newblock ACM.

\bibitem[\protect\citeauthoryear{O'Mahony and Shmoys}{2015}]{o2015data}
O'Mahony, E., and Shmoys, D.~B.
\newblock 2015.
\newblock Data analysis and optimization for (citi) bike sharing.
\newblock In {\em AAAI},  687--694.

\bibitem[\protect\citeauthoryear{Shaheen, Guzman, and
  Zhang}{2010}]{shaheen2010bikesharing}
Shaheen, S.; Guzman, S.; and Zhang, H.
\newblock 2010.
\newblock Bikesharing in europe, the americas, and asia: past, present, and
  future.
\newblock {\em Transportation Research Record: Journal of the Transportation
  Research Board} (2143):159--167.

\bibitem[\protect\citeauthoryear{Silver \bgroup et al\mbox.\egroup
  }{2014}]{silver2014deterministic}
Silver, D.; Lever, G.; Heess, N.; Degris, T.; Wierstra, D.; and Riedmiller, M.
\newblock 2014.
\newblock Deterministic policy gradient algorithms.
\newblock In {\em Proceedings of the 31st International Conference on Machine
  Learning (ICML-14)},  387--395.

\bibitem[\protect\citeauthoryear{Singla \bgroup et al\mbox.\egroup
  }{2015}]{singla2015incentivizing}
Singla, A.; Santoni, M.; Bart{\'o}k, G.; Mukerji, P.; Meenen, M.; and Krause,
  A.
\newblock 2015.
\newblock Incentivizing users for balancing bike sharing systems.
\newblock In {\em AAAI},  723--729.

\bibitem[\protect\citeauthoryear{Tang}{2017}]{tang2017reinforcement}
Tang, P.
\newblock 2017.
\newblock Reinforcement mechanism design.
\newblock In {\em Proceedings of the Twenty-Sixth International Joint
  Conference on Artificial Intelligence, IJCAI-17},  5146--5150.

\bibitem[\protect\citeauthoryear{Van~Seijen \bgroup et al\mbox.\egroup
  }{2017}]{van2017hybrid}
Van~Seijen, H.; Fatemi, M.; Romoff, J.; Laroche, R.; Barnes, T.; and Tsang, J.
\newblock 2017.
\newblock Hybrid reward architecture for reinforcement learning.
\newblock In {\em Advances in Neural Information Processing Systems},
  5392--5402.

\bibitem[\protect\citeauthoryear{Xue \bgroup et al\mbox.\egroup
  }{2016}]{xue2016avicaching}
Xue, Y.; Davies, I.; Fink, D.; Wood, C.; and Gomes, C.~P.
\newblock 2016.
\newblock Avicaching: A two stage game for bias reduction in citizen science.
\newblock In {\em Proceedings of the 2016 International Conference on
  Autonomous Agents \& Multiagent Systems},  776--785.
\newblock International Foundation for Autonomous Agents and Multiagent
  Systems.

\bibitem[\protect\citeauthoryear{Yang \bgroup et al\mbox.\egroup
  }{2016}]{yang2016mobility}
Yang, Z.; Hu, J.; Shu, Y.; Cheng, P.; Chen, J.; and Moscibroda, T.
\newblock 2016.
\newblock Mobility modeling and prediction in bike-sharing systems.
\newblock In {\em Proceedings of the 14th Annual International Conference on
  Mobile Systems, Applications, and Services},  165--178.
\newblock ACM.

\end{thebibliography}

\end{document}